\documentclass[10pt,twocolumn,letterpaper]{article}

\usepackage{tabularray}
\usepackage{pifont}
\usepackage{multirow}
\usepackage{arydshln}
\usepackage{makecell}
\usepackage{booktabs}
\usepackage{graphicx}
\usepackage[pagenumbers]{cvpr} 
\usepackage[pagebackref,breaklinks,colorlinks,citecolor=cvprblue]{hyperref}
\usepackage[T1]{fontenc}

\usepackage[dvipsnames]{xcolor}

\definecolor{cvprblue}{rgb}{0.21,0.49,0.74}

\title{Gaussian Harmony: Attaining Fairness in Diffusion-based Face Generation Models}

\author{
  Basudha Pal$^{1}$\thanks{Indicates equal contribution.} \and
  Arunkumar Kannan$^1$\footnotemark[1]\and
  Ram Prabhakar Kathirvel$^{1}$ \and
  Alice J. O'Toole$^{2}$ \and
  Rama Chellappa$^{1}$\\
  $^1$Johns Hopkins University, $^2$The University of Texas at Dallas\\
  {\tt\small \{bpal5, akannan7, rprabha3, rchella4\}@jhu.edu}, \tt\small otoole@utdallas.edu
}

\begin{document}
\maketitle
\begin{abstract}
Diffusion models have achieved great progress in face generation. However, these models amplify the bias in the generation process, leading to an imbalance in distribution of sensitive attributes such as age, gender and race. This paper proposes a novel solution to this problem by balancing the facial attributes of the generated images. We mitigate the bias by localizing the means of the facial attributes in the latent space of the diffusion model using Gaussian mixture models (GMM). Our motivation for choosing GMMs over other clustering frameworks comes from the flexible latent structure of diffusion model. Since each sampling step in diffusion models follows a Gaussian distribution, we show that fitting a GMM model helps us to localize the subspace responsible for generating a specific attribute. Furthermore, our method does not require retraining, we instead localize the subspace on-the-fly and mitigate the bias for generating a fair dataset. We evaluate our approach on multiple face attribute datasets to demonstrate the effectiveness of our approach. Our results demonstrate that our approach leads to a more fair data generation in terms of representational fairness while preserving the quality of generated samples. 
\end{abstract}
    
\section{Introduction}
\label{sec:intro}

Generative models have revolutionized image generation tasks by enabling deep learning networks to generate data that closely resemble real-world examples \cite{r1}, \cite{r2}, \cite{r3}, \cite{r4}, \cite{r5}. These models have a wide range of applications across various domains, including image synthesis \cite{r6}, text generation \cite{r7}, and data augmentation \cite{r8}. However, increased utilization of generative models has raised concerns regarding the potential biases associated with these models.

\begin{figure} [h]
    \centering
    \includegraphics[width = \linewidth]{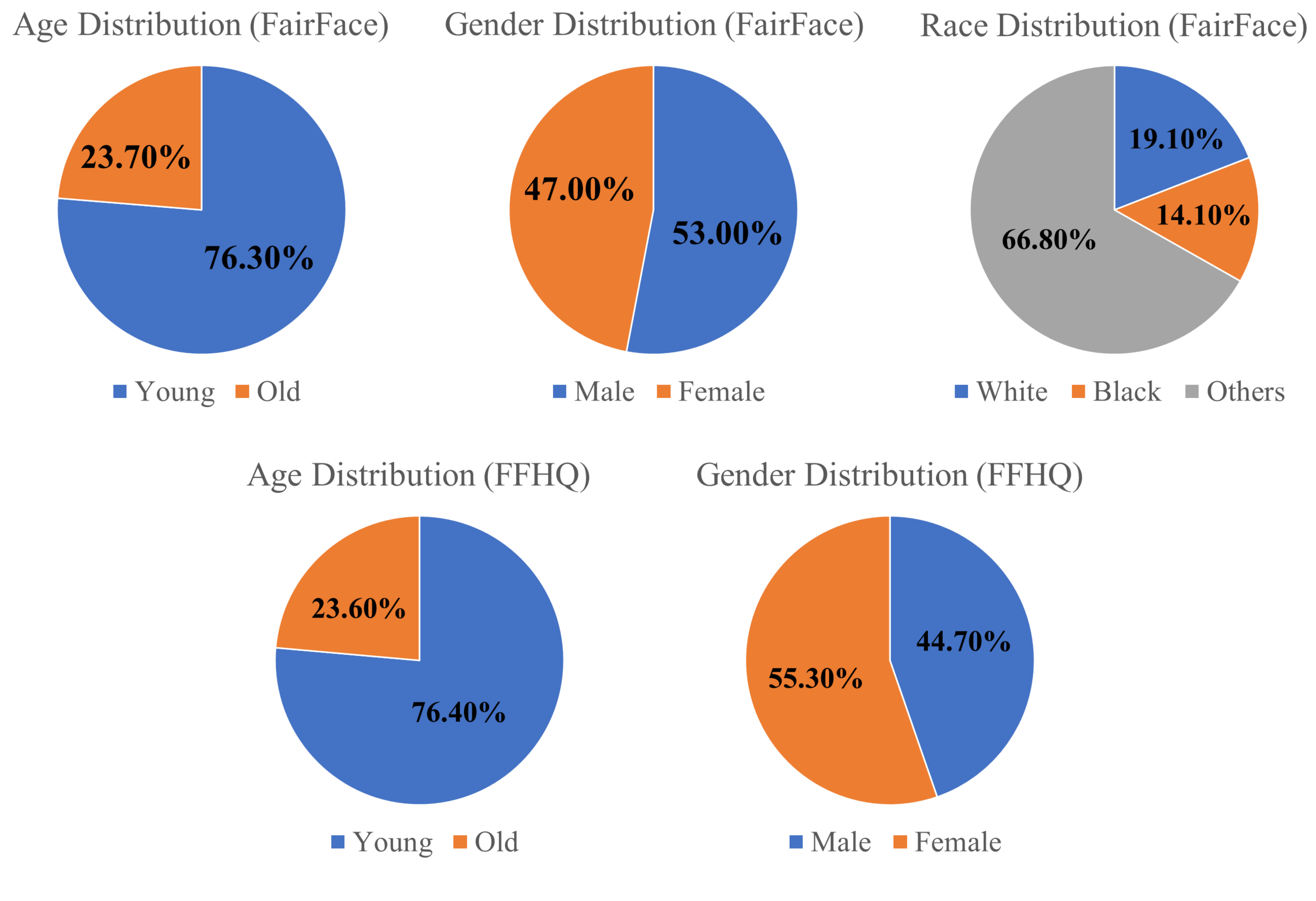}
    \caption{Data distribution in the original unbalanced datasets. \textbf{Top}: Distribution of FairFace dataset with respect to age, gender and race, \textbf{Bottom}: Distribution of FFHQ dataset with respect to age and gender.}
    \label{pie_chart}
\end{figure}

 Prior work exists in mitigating bias in face recognition and verification tasks as demonstrated in \cite{r9} and \cite{r10}. Recently, there has been a growing interest in the topic of equity within generative models, as highlighted by the works of \cite{r11}, \cite{r12}, \cite{r13}. Fairness in this context refers to achieving equality in the representation of specific sensitive attributes such as age, gender, and race. For instance, a generative model that maintains an equal likelihood of generating both male and female samples can be considered fair in terms of the gender attribute \cite{r14}, \cite{r15}. Generative models have even been employed in high-stakes domains such as suspect facial profiling \cite{r16}. In such instances, if these models exhibit biases concerning sensitive attributes such as gender or race, there is a risk of wrongly incriminating individuals. In addition, generative models have been used to generate data for training downstream models, such as skin-lesion diagnosis classifiers \cite{r17}. The biases in generative models can propagate to downstream models, compounding  fairness concerns in the overall system.

Given these  challenges and the recent attention to fairness in generative models, it is crucial to ensure equitable representation and mitigate potential biases in these systems. There is much work in bias mitigation in generative models in recent years \cite{r18}, \cite{r19}. However, these studies involve  mitigating bias in Generative Adversarial Networks (GANs), which amplify and introduce bias in the generated data distribution. Teo and Abdollahzadeh et al. \cite{r15} proposed a transfer learning-based approach for fair image generation to sample an equal number of images from each of the sensitive attributes. Recently, Diffusion models \cite{r20} have gained  attention for their ability to outperform GANs in terms of image generation quality and their
ability to address  some of GANs' limitations such as mode collapse and unstable training convergence. Diffusion models are being used for various  computer vision applications as shown  in \cite{r21}. Yet, it has been observed that diffusion models are not immune to bias issues, highlighting the continued importance of research into mitigating biases in generative models. Recently, \cite{r22} showed that diffusion models introduce and amplify bias in face generation tasks \cite{r22}.

\begin{figure*}
    \centering
    \includegraphics[width = 0.8\linewidth]{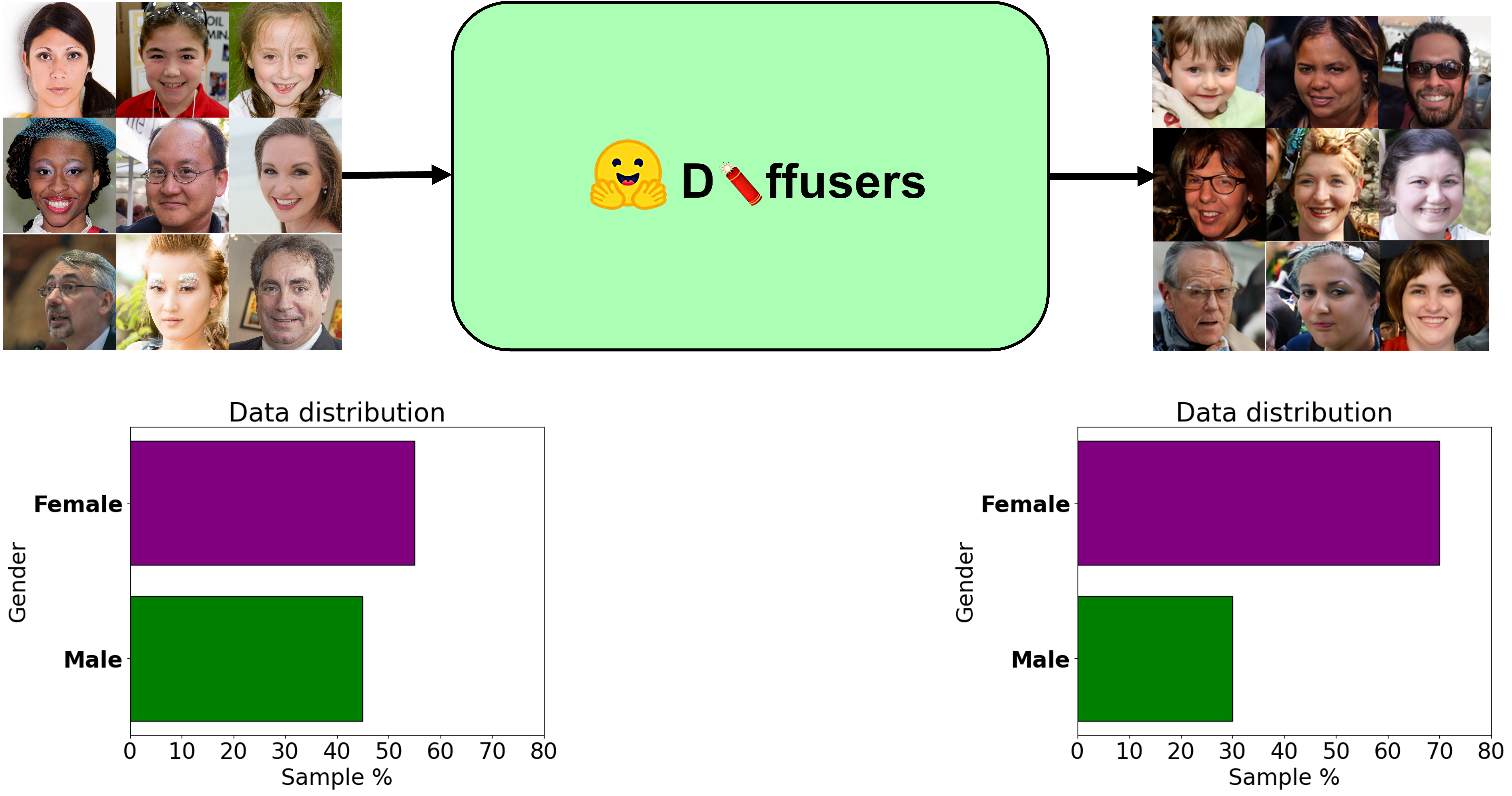}
    \caption{Bias amplification by diffusion models in face generation (Binary Gender attribute - Male/Female)}
    \label{imbalanced}
\end{figure*}

\begin{figure*}
    \centering
    \includegraphics[width = 0.8\linewidth]{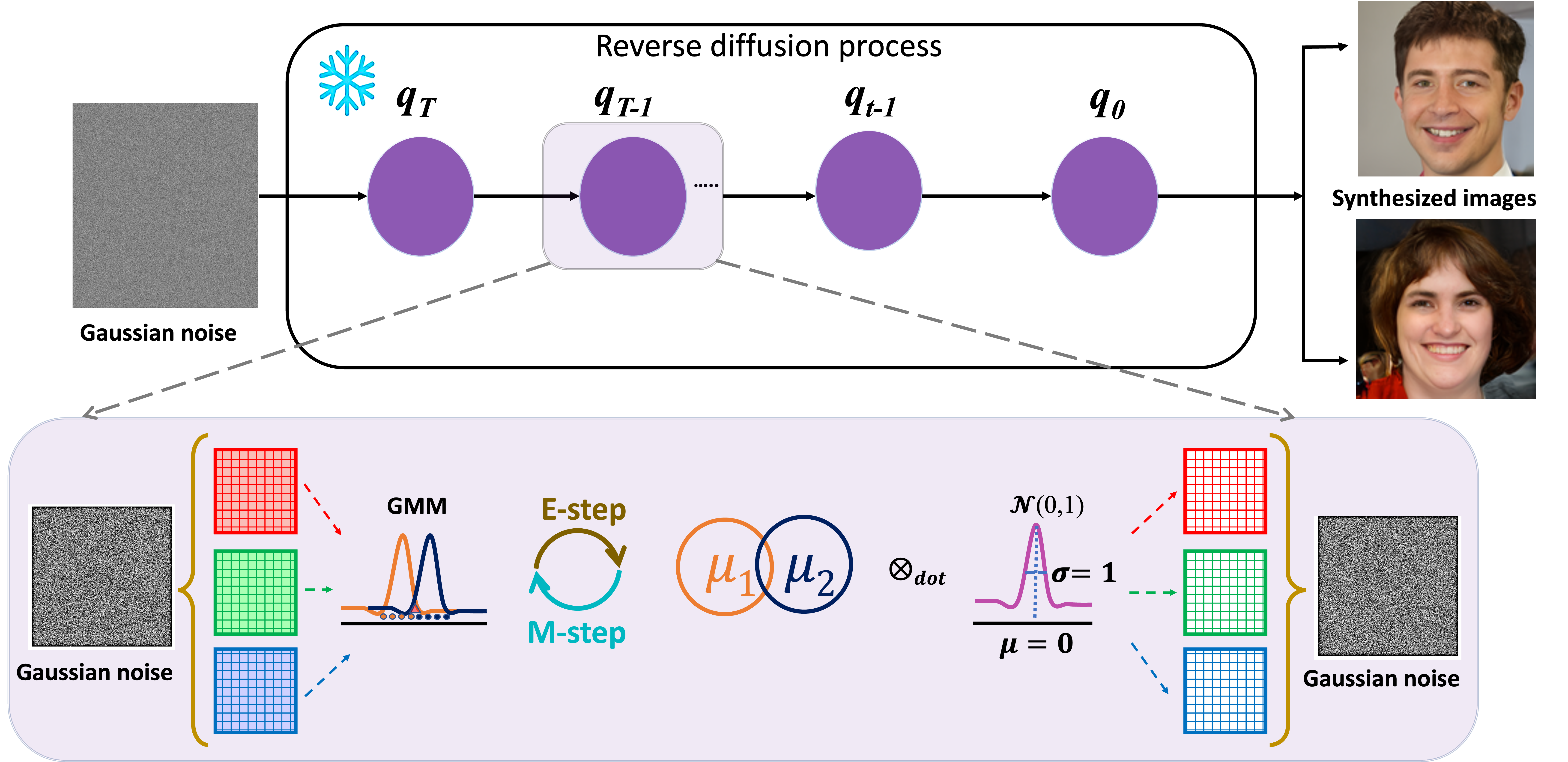}
    \caption{Illustration of our method to mitigate the bias in diffusion model. We use a GMM to fit the means of respective attribute classes for each channel in the reverse diffusion process. We concatenate the resultant means to obtain the ``bias-corrected" noise which we utilize to generate the image with a particular attribute class.}
    \label{workflow}
\end{figure*}

In this paper, we build on previous work suggesting that diffusion models amplify bias while generating images. Therefore, our main focus lies in mitigating the bias introduced/amplified by diffusion models when trained with balanced or imbalanced training data. We follow the same initial setup as Perera et al. \cite{r22} and take it a step further by proposing a bias mitigation technique. In our approach, we fit Gaussian Mixture Models (GMMs) to localize the means of the various classes of sensitive attributes in the latent space during the reverse diffusion process, thereby reducing bias without retraining. We perform this analysis in the case of balanced and imbalanced input for training the diffusion model. The major contributions of our paper are:
\begin{itemize}
\item We conduct experiments to show that the bias in training data is amplified by the diffusion model.
\item We report an initial baseline for bias mitigation in diffusion models for face generation across three sensitive attributes - age, gender and race. 
\item We achieve bias mitigation without retraining. 
\end{itemize}

\section{Related Work}
\vspace{-5pt}
\textbf{Face generative models:} 
In recent times, face image synthesis has been significantly transformed by the power of generative models. This revolution began with generative adversarial networks (GANs) \cite{r23} where a generator and a discriminator, are trained simultaneously. Through iterative training, GANs learn to map complex patterns from the latent space to produce realistic data, and were previously considered state-of-the-art in various applications \cite{r24}, \cite{r25}, \cite{r26} including synthetic face generation \cite{r27}, {\cite{r28}. However, GANs are often difficult to train and suffer from the problem of mode collapse. The more recent denoising diffusion probabilistic models (DDPM) are parameterized as a Markov chain whose transition distributions are Gaussian. DDPMs consists of a forward process and a reverse process. In the forward process, data undergoes sequential transformation into isotropic Gaussian noise through the addition of Gaussian noise in $T$ diffusion time steps. During evaluation, the reverse diffusion process initiates from an isotropic Gaussian state and iteratively removes noise to produce a generative sample aligned with the training data distribution. These models provide high quality results with a simpler objective and stable convergence, thereby proving to be better than GANs in multiple applications \cite{r29}, \cite{r30} including face generation \cite{r31}, \cite{r32}. As DDPMs are becoming widespread in facial image generation, we utilize these models to explore the fundamental problem of bias for this task.
\\
\textbf{Bias in computer vision:} 
In the field of computer vision, bias has been a long standing problem due to its prevalence in various applications. For example, \cite{r33} investigated how face recognition accuracy is influenced by race and skin tone, whereas other research \cite{r34}, \cite{r35}, \cite{r36} has examined the performance of face recognition systems across various factors including skin brightness, hairstyles and use of cosmetics. In face recognition and verification tasks, Albeiro et al. \cite{r37} explored the effect of gender balance in training data and concluded that balancing data for training in terms of gender does not necessarily lead to a gender-balanced output in testing. 
\\
\textbf{Bias in face generative models:} 
Since face generative models have gained popularity, several studies have shown that models like GANs and DDPMs are susceptible to bias. Jain et al. \cite{r38} revealed that popular face generative models based on GANs amplify biases associated with the distribution of gender and skin tone of individuals in training data, while Maluleke et al. \cite{r39} conducted a study on how GANs are affected by the racial composition of training data. In the more recent diffusion models, Luccioni et al. \cite{r40} introduced a novel approach to investigate and measure social biases in text-to-image systems and showed that across all target attributes, text-to-image systems demonstrate a notable over-representation of White individuals and males. Naik et al. \cite{r41} examined how text-to-image generative models depict bias in profession, personality and other usual situations in their output with respect to attributes such as gender, age, race and location. Recently, \cite{r22} explored the bias in diffusion models for unconditional face generation. They extensively studied the effect of balanced and imbalanced training data on unconditional diffusion models for face generation. Their findings suggest that unconditional diffusion model based face generators amplify the distribution bias in the training data in terms of age, gender and race across both balanced and imbalanced datasets.
\\
\textbf{Bias mitigation techniques in face generative models:} In GANs for face generation, multiple approaches have been used to mitigate the inherent bias \cite{r18}, \cite{r19}. Grover et al. \cite{r43} correct bias in deep generative models using likelihood-free importance weighting, employing a trained probabilistic classifier to estimate the unknown likelihood ratio between model and true distributions. Choi et al. \cite{r44} proposed a method that tackles dataset bias in deep generative models by leveraging weak supervision from a small, unlabeled reference dataset, enabling efficient learning and generating test data closely aligned with the reference dataset distribution. Researchers in \cite{r15} achieve this using a simpler transfer learning based approach.
However, these approaches are limited to reducing bias in GANs and  require retraining. To the best of our knowledge, no prior research has investigated the mitigation of bias in diffusion models concerning face generation without retraining.
                                                
\section{Proposed Method}
\vspace{-5pt}

Our approach centers on developing a diffusion model to generate images unconditionally, with a focus on generating facial images. Leveraging this generative framework, we aim to target the specific subspace in the latent space of the diffusion model responsible for the generation of a predefined attribute. To achieve this localization, we employ a GMM, designed to unravel the intricate patterns within the latent space and identify the underlying subspace contributing to the synthesis of the targeted facial features. This dual-phase approach not only facilitates the creation of realistic facial images through diffusion modeling, but also enables the extraction of attribute-specific information using Gaussian mixture modeling.

\subsection{Problem Setup}
\vspace{-5pt}
Our dataset comprises facial images annotated with specific attributes, including age ($a$), gender ($g$), and race ($r$). Each facial image is denoted as $x_i$, where $i$ represents the index of the image in the dataset. The corresponding attribute labels associated with each image are represented by $a_i$, $g_i$ and $r_i$ indicating the age, gender, and race labels, respectively. Our goal is to use diffusion models for unconditional image generation, where the generative process is modeled as $x_i=G\left(Z_i, \theta\right)$, with $Z_i$ being the latent representation corresponding to the $i$-th facial image, and $\theta$ representing the parameters of the diffusion model.

\subsection{Stage 1: Unconditional Image Generation}
\vspace{-5pt}
Diffusion models are a class of generative models that model the data distribution in the form of $p_\theta\left(\mathbf{x}_0\right):=$ $\int p_\theta\left(\mathbf{x}_{0: T}\right) d \mathbf{x}_{1: T}$. The diffusion process (a.k.a. forward process) gradually adds Gaussian noise to the data and eventually corrupts the data $\mathbf{x}_0$ into an approximately pure Gaussian noise $\mathbf{x}_T$ using a variance schedule $\beta_1, \ldots, \beta_T$ :
$$
\begin{aligned}
& q\left(\mathbf{x}_{1: T} \mid \mathbf{x}_0\right):=\prod_{t=1}^T q\left(\mathbf{x}_t \mid \mathbf{x}_{t-1}\right) \\
& q\left(\mathbf{x}_t \mid \mathbf{x}_{t-1}\right):=\mathcal{N}\left(\mathbf{x}_t ; \sqrt{1-\beta_t} \mathbf{x}_{t-1}, \beta_t \mathbf{I}\right)
\end{aligned}
$$

Reversing the forward process allows the sampling of new data $\mathbf{x}_0$ by starting from $p\left(\mathbf{x}_T\right)=\mathcal{N}\left(\mathbf{x}_T ; \mathbf{0}, \mathbf{I}\right)$. The reverse process is defined as a Markov chain where each step is a learned Gaussian transition $\left(\boldsymbol{\mu}_\theta, \boldsymbol{\Sigma}_\theta\right)$ :
$$
\begin{aligned}
& p_\theta\left(\mathbf{x}_{0: T}\right):=p\left(\mathbf{x}_T\right) \prod_{t=1}^T p_\theta\left(\mathbf{x}_{t-1} \mid \mathbf{x}_t\right), \\
& p_\theta\left(\mathbf{x}_{t-1} \mid \mathbf{x}_t\right):=\mathcal{N}\left(\mathbf{x}_{t-1} ; \boldsymbol{\mu}_\theta\left(\mathbf{x}_t, t\right), \boldsymbol{\Sigma}_\theta\left(\mathbf{x}_t, t\right)\right) .
\end{aligned}
$$

Training diffusion models relies on minimizing the variational bound on $p(x)$ 's negative log-likelihood. The commonly used optimization objective $L_{\mathrm{DM}}$ [23] reparameterizes the learnable Gaussian transition as $\epsilon_\theta(\cdot)$, and temporally reweights the variational bound to trade for better sample quality:
$$
L_{\mathrm{DM}}(\theta):=\mathbb{E}_{t, \mathbf{x}_0, \boldsymbol{\epsilon} \sim \mathcal{N}(\mathbf{0}, \mathbf{I})}\left[\left\|\boldsymbol{\epsilon}-\boldsymbol{\epsilon}_\theta\left(\mathbf{x}_t, t\right)\right\|^2\right],
$$
where $\mathbf{x}_t$ can be directly approximated by $\mathbf{x}_t=\sqrt{\bar{\alpha}_t} \mathbf{x}_0+$ $\sqrt{1-\bar{\alpha}_t} \epsilon$, with $\bar{\alpha}_t:=\prod_{s=1}^t \alpha_s$ and $\alpha_t:=1-\beta_t$.

To sample data $\mathbf{x}_0$ from a trained diffusion model $\epsilon_\theta(\cdot)$, we iteratively denoise $\mathbf{x}_t$ from $t=T$ to $t=1$ with noise $\mathbf{z}$

$\mathbf{x}_{t-1}=\frac{1}{\sqrt{\alpha_t}}\left(\mathbf{x}_t-\frac{1-\alpha_t}{\sqrt{1-\bar{\alpha}_t}} \boldsymbol{\epsilon}_\theta\left(\mathbf{x}_t, t\right)\right)+\sigma_t \mathbf{z}$

\subsection{Stage 2: Gaussian Mixture Models - a clustering perspective}
\vspace{-5pt}
Our main objective is to discern distinct clusters within the latent space that correspond to different attribute values, such as age, gender, and race. We achieve this using GMM where we strategically utilize the latent codes generated using diffusion model during each time step in the forward Markov process. These latent representations encapsulate the high-dimensional features that contribute to the synthesis of face images with specific attributes. 

The GMM is a probabilistic model that assumes that the data is generated from a mixture of several Gaussian distributions. We utilize this assumption because of the Gaussian nature of generated latent codes. The latent space representations of the generated face images were considered as the input data for the GMM.

Let $\mathbf{Z} \in \mathbb{R}^{N \times d}$  represent the latent space of the diffusion model, where $N$ is the number of generated samples, and $d$ is the dimensionality of the latent space at time step $t$. Each row of $\mathbf{Z}$ corresponds to a latent representation of a face image. We train a GMM to model the distribution of $\mathbf{Z}$.

$p(\mathbf{Z};\boldsymbol{\theta})=\sum_{i=1}^K \pi_i \mathcal{N}\left(\mathbf{Z}, \boldsymbol{\mu}_i, \boldsymbol{\Sigma}_i\right)$

where $0 \leqslant \pi_i \leqslant 1, \quad \sum_{i=1}^K \pi_i=1$ and $\boldsymbol{\theta}=\left\{\boldsymbol{\mu}_i, \boldsymbol{\Sigma}_i, \boldsymbol{\pi}_i\right\}_{i=1}^K$. $K$ is the number of components in the GMM, $\pi_i$ is the weight of the $i$-th component, $\mathcal{N}\left(\cdot ; \boldsymbol{\mu}_i, \boldsymbol{\Sigma}_i\right)$ is a multivariate Gaussian distribution modeling the latent space. 

The main objective of GMM is to determine $\boldsymbol{\mu}_i$ and $\boldsymbol{\Sigma}_i$ associated with each Gaussian mixtures. We perform this using expectation-maximization (EM) algorithm. The Expectation Maximization (EM) algorithm is an iterative method for finding the maximum likelihood estimates of $\boldsymbol{\mu}_i$ and $\boldsymbol{\Sigma}_i$. The algorithm consists of two steps: the E-step in which a function for the expectation of the log-likelihood is computed based on the current parameters, and an M-step where the parameters found in the first step are maximized. Every EM iteration increases the log-likelihood function. For convergence, we can check the log-likelihood and stop the algorithm when a certain threshold $\epsilon$ is reached, or alternatively when a predefined number of steps is reached. The update equations in E-step and M-step can be summarized as follows:
$
\begin{aligned}
    \quad \tau_n^k=\frac{\pi_k P_k\left(y_n ; \theta_k\right)}{\sum_{k=1}^k \pi_k P_k\left(y_n ; \theta_k\right)},
    \boldsymbol{\mu}_k=\frac{1}{N_k} \sum_{n=1}^N r_{n k} \boldsymbol{x}_n
\end{aligned}
$

$
\begin{aligned}
        \quad \boldsymbol{\Sigma}_k=\frac{1}{N_k} \sum_{n=1}^N \tau_n^k \left(\boldsymbol{x}_n-\boldsymbol{\mu}_k\right)\left(\boldsymbol{x}_n-\boldsymbol{\mu}_k\right)^{\top},
    \pi_k=\frac{N_k}{N}
\end{aligned}$

The update process will repeat until algorithm convergence, typically achieved when the model parameters do not change significantly from one iteration to the next.

\subsection{Stage 3: Localization of specific attributes}
\vspace{-5pt}
To disentangle the latent components of the diffusion model corresponding to different attributes, namely age, gender, and race, we leverage the expressivity hierarchy assumption: age $>$ gender $>$ race. This assumption is derived from prior research indicating that latent features exhibit varying degrees of expressivity concerning these attributes \cite{r45}. To operationalize this assumption, we calculate the Kullback-Leibler (KL) divergence between GMM components.

We assume that the latent components contributing to age-related variations have higher expressivity than those corresponding to gender and race. Similarly, the components responsible for gender-related variations have higher expressivity than those for race. For each pair of GMM components, we compute the KL divergence based on their multivariate Gaussian distributions in the latent space. Specifically, let $P_i$ and $P_j$ be two components, then $D_{\mathrm{KL}}\left(P_i \| P_j\right)$ denotes the KL divergence from $P_i$ to $P_j$. Higher KL divergence values suggest greater dissimilarity between components.

We assign attributes to the latent components based on the computed KL divergences. Specifically, if $D_{\mathrm{KL}}($ age $\|$ gender $)$ is larger than $D_{\mathrm{KL}}($ gender $\|$ race $)$, and $D_{\mathrm{KL}}$ (age $\|$ race) is the largest, we interpret the corresponding components as primarily associated with age. We follow a similar procedure for gender and race.

\section{Experiments}
\vspace{-5pt}
In this section, we present experiments demonstrating that unconditional diffusion models accentuate the existing bias in the dataset with respect to age, gender and race. We show the effectiveness of our method in reducing bias by fitting a GMM in the reverse diffusion process. 
\subsection{Datasets}
\vspace{-5pt}
We verify our experiments on two popular datasets for exploring bias in face genratation.
\begin{itemize}
    \item \textbf{FFHQ Dataset:} The Flickr Faces HQ (FFHQ Dataset) is a commonly utilized dataset released by NVIDIA for tasks involving face generation \cite{r47}. This dataset comprises 70,000 aligned and cropped images with a high resolution of 1024 × 1024 pixels, obtained from Flickr. Using the FFHQ dataset, we assess potential biases in face generation models related to gender and age attributes. 
    \item \textbf{FairFace Dataset:} The FairFace dataset,introduced in \cite{r46} consists of 108,501 aligned and cropped facial images. This dataset spans seven racial/ethnic groups, a wide range of ages from 0 to above 70 and encompasses both genders. It specifically focuses on images of non-public figures to minimize selection bias, aiming to improve fairness and accuracy in face recognition systems by providing a more diverse dataset.
\end{itemize}

We limit our analysis to binary classes for age and gender and three classes for race. In FFHQ, the ages of the individuals are provided, which we categorize those below 40 as `Young' and those above 40 as `Old'. In FairFace, there are nine categories of age attributes which we reorganize into two groups, (i) Young (`0-2' $\cup$ `3-9' $\cup$ `10-19' $\cup$ `20-29' $\cup$ `30-39'), (ii) Old (`40-49' $\cup$ `50-59 $\cup$ `60-69' $\cup$ `more than 70'). Further, FairFace has seven categories of the race attribute which we reorganize into three groups (i) White, (ii) Black and (iii) Other (`Indian' $\cup$ `East Asian' $\cup$ `Southeast Asian' $\cup$ `Middle Eastern'$\cup$ `Latino'). Figure \ref{pie_chart} illustrates the distribution of data in FFHQ and FairFace datasets after our manipulations. 

\subsection{Attribute classifier}
\vspace{-5pt}
We train an off-the-shelf attribute classifier ($\mathcal{A}$) to predict the attributes of generated face images in evaluation phase. We train $\mathcal{A}$ with the training sets of both FFHQ and FairFace datasets. $\mathcal{A}$ is based on a standard multi-head ResNet101, which classifies $\textit{r}_i$ into one of the three categories and $\textit{a}_i$ and $\textit{g}_i$ into one of the two categories. After training $\mathcal{A}$, we use it for evaluation on generated images and validation images of both datasets.

We realize that in our experiments, $\mathcal{A}$ can introduce additional bias, therefore, we attempt to keep the classifier constant throughout our analyses. In order to keep the bias of $\mathcal{A}$ constant, we train it jointly for both datasets and use it as a common attribute classifier across both datasets. We pass the validation set, $\mathcal{V}$, through $\mathcal{A}$ for each dataset individually and obtain the counts for classes of each $\textit{a}_i$, $\textit{g}_i$, and $\textit{r}_i$ where $i=0,1$ for $\textit{a}$ and $\textit{g}$ and $i = 0,1,2$ for $\textit{r}$. We further strengthen our claim that the bias of $\mathcal{A}$ is kept constant by making all our comparisons with these data counts.

\subsection{Unconditional face generation}
\vspace{-5pt}
We conduct experiments separately for FFHQ and FairFace datasets. We train an unconditional diffusion model, $\mathcal{D}$ with the help of Hugging Face Diffusers using the $x_i's$ and generate as many images as in the validation set of the corresponding dataset. We then pass these through $\mathcal{A}$. It is observed that the bias is amplified for each of the attributes when compared to the counts in the validation data that was initially passed through the attribute classifier as shown in Figure \ref{imbalanced}. This experiment is conducted for both the FFHQ and FairFace datasets.  In FFHQ, the analysis is done only for $\textit{a}_i$ and $\textit{g}_i$, while in FairFace, it is done for all $\textit{a}_i$, $\textit{g}_i$, and $\textit{r}_i$.
\subsection{Bias corrected diffusion model}
\vspace{-5pt}
The final stage in our pipeline is to correct this bias in each of the attributes that is introduced or amplified by $\mathcal{D}$. During the reverse diffusion process (evaluation), instead of starting from isotropic Gaussian noise and reaching the generated image, we localize the means of the $\textit{z}_i's$ which correspond to the specific classes of attributes by fitting a GMM as shown in Figure. \ref{workflow}. The GMM seperates out the different components of the specific attribute and we fit those means to generate an equal number of images from each class. In this manner, we are able to mitigate bias in the generated image set. Once the generated images are obtained, we pass them through $\mathcal{A}$ and compare the counts of each class of attributes to their uncorrected counterpart (as in the previous subsection). We are able to correct the bias in terms of image count per class by separating out the components and equally sampling from each of the distributions.
\subsection{Implementation Details}
\vspace{-5pt}
We use an off-the-shelf multi-head classifier with a ResNet-101 backbone to predict the classes in age, gender and race. We optimize the training with the Adam optimizer and a learning rate of $10^{-4}$. For the diffusion models we utilize the Hugging Face diffusers in order to train both FFHQ and FairFace. For the entire unbalanced dataset, we train for 30 epochs using the Adam optimizer with a learning rate of $10^{-4}$. In case of the smaller, balanced datasets that we curate, we increase the number of epochs to 50 for each balanced attribute. For the GMM model, we fit two components for gender and age and three for race to separate out our classes. We then use these means as the means of the noise tensors at the 350$^{th}$ timestep of the reverse diffusion process. We set the covariance type in GMM model to be `diagonal'. We generate an equal number of images from the bias corrected diffusion model as in the validation set that was passed through $\mathcal{A}$. Further we compare the number of classes of each attribute in that validation set and the corrected diffusion model after passing both through $\mathcal{A}$.

\begin{table*}
\centering
\resizebox{\linewidth}{!}{%
\begin{tabular}{c||c||c|c|c|c} 
\toprule
Attribute               & Class  & FFHQ - Balanced            & FFHQ - Imbalanced          & FairFace - Balanced & FairFace - Imbalanced  \\ 
\bottomrule
\multirow{2}{*}{Gender} & Female & 19.02                      & 14.02                      & 22.47               & 19.78                  \\
                        & Male   & 21.82                      & 17.85                      & 23.50               & 19.51                  \\ 
\bottomrule
\multirow{2}{*}{Age}    & Old    & 23.62                      & 23.68                      & 28.54               & 25.57                  \\
                        & Young  & 24.19                      & 14.51                      & 23.04               & 17.39                  \\ 
\bottomrule
\multirow{3}{*}{Race}   & Black  & \ding{55} & \ding{55} & 30.59               & 24.98                  \\
                        & White  & \ding{55} & \ding{55} & 28.06               & 24.19                  \\
                        & Others & \ding{55} & \ding{55} & 25.01               & 18.26                  \\
\bottomrule
\end{tabular}
}
\caption{FID scores of the images generated by diffusion models trained on balanced and imbalanced data.}
\label{table1}
\end{table*}

\begin{figure*}
    \centering
    \includegraphics[width = \linewidth]{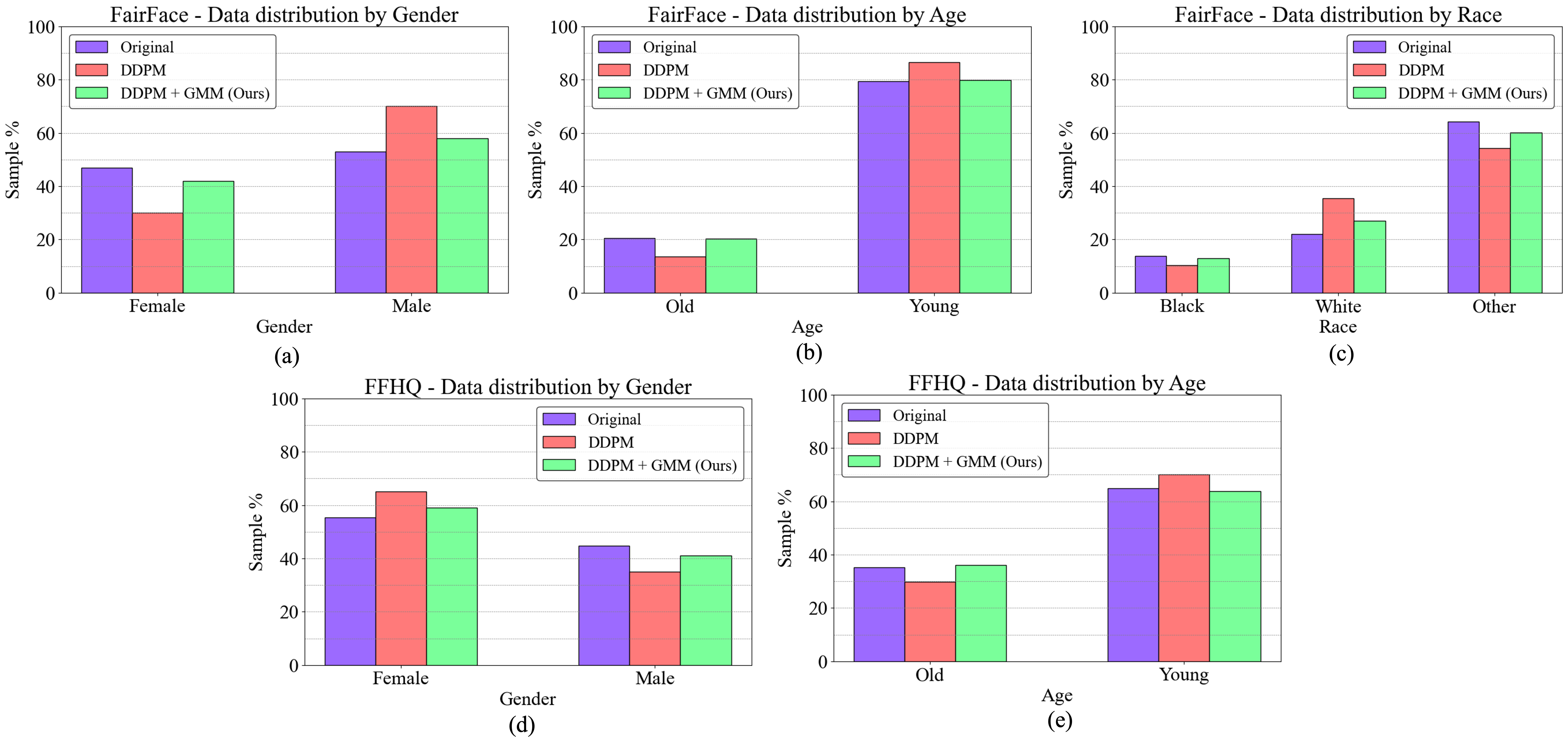}
    \caption{Results of our experiments on original  data distribution (in purple) of each class of attribute in percentage. When the diffusion model is trained on the original dataset (biased), the unconditionally generated images show an amplification in the bias for a particular class (in red). Using our approach, we demonstrate the reduction in bias in unconditionally generated images that was introduced by the diffusion model (in green).}
    \label{imbalanced}
\end{figure*}

\section{Results and Discussion}
\vspace{-5pt}
In all our experiments, we note a consistent improvement in fairness when applying GMM to separate out noisy latent codes of the generated samples into different classes of one attribute. As this technique does not require any retraining and is just an additional step during evaluation for any diffusion model, we anticipate its advantageous impact on a wide range of tasks in future research endeavors. 

The first set of results is on the quality of the unconditional diffusion model that we trained. The Frechet inception distance (FID)  is a commonly utilized metric to assess the perceptual quality of images produced by a generative model. FID gauges image quality by comparing feature distributions extracted from the Inception network between training and generated images. Thus, lower FID scores signify higher perceptual quality. We present the FID scores for generated samples in different attribute classes of gender, race, and age in Table \ref{table1}. Overall, we can see that the FID scores for attribute classes in images produced by diffusion models trained with imbalanced data are lower relative to the balanced data. This is associated with the more number of images in the imbalanced data.  

In our experiments, we establish that unconditional diffusion models when used for face generation, create or amplify inherent biases. Figure \ref{imbalanced} is the best visual illustration of this. We quantify bias in face generative models by the number of images of each category generated.
As shown in Figure \ref{imbalanced}(a), the diffusion model, when trained on the complete FairFace dataset, produces a higher percentage of male images compared to the gender distribution in the training data. This implies a bias towards generating male face images, potentially amplifying training data bias. 

Conversely, Figure \ref{imbalanced}(d) reveals that utilizing the diffusion model trained on the FFHQ dataset results in a bias towards female images, emphasizing that the direction of bias depends on the gender distribution in the training dataset. However, our method intervenes as a corrective measure, mitigating this bias by localizing means and incorporating them into the reverse diffusion process. As seen in the case of FairFace, the diffusion model amplified the existing bias by generating more male images but when we use our technique we are able to reduce the number of male images generated and increase the number of female images generated. Similarly, for FFHQ when we sample from the separated distributions, we are able to reduce the bias of the diffusion model by generating more male images than female images. 
Examining Figure \ref{imbalanced}(b), the age distribution of the generated data from the diffusion model trained on the complete FairFace dataset indicates that, the proportion of faces in age groups above 40 is lower than the training distribution. In the FFHQ training data there was existing bias towards young faces but the trained diffusion model amplifies it. 

A similar trend is observed in the case of the FFHQ  dataset in Figure \ref{imbalanced}(e). This suggests that the FFHQ-trained diffusion model also has a bias towards generating younger face images.  Our method serves as a corrective measure, addressing this age-related bias using a GMM based approach in the reverse diffusion process. By applying our technique to FairFace, which experienced an amplification of the existing bias in generating younger faces, we demonstrate a reduction in the number of younger faces generated and an increase in the representation of older age groups. Similarly, when our method is applied to FFHQ, it also generates more old faces and helps mitigate the bias of the diffusion model, resulting in a more balanced generation of age-diverse facial images. This underscores the effectiveness of our approach in counteracting and mitigating age-related biases present in the generated data.
Figure \ref{imbalanced}(c) demonstrates that the diffusion model trained on the FairFace dataset produces a higher proportion of face images belonging to the white racial class, indicating a potential perpetuation of bias towards this specific group. We correct the  bias using our method and are capable of obtaining a distribution aligned to the training data thereby mitigating the bias introduced by the diffusion models.

\begin{figure*}
    \centering
    \includegraphics[width = \linewidth]{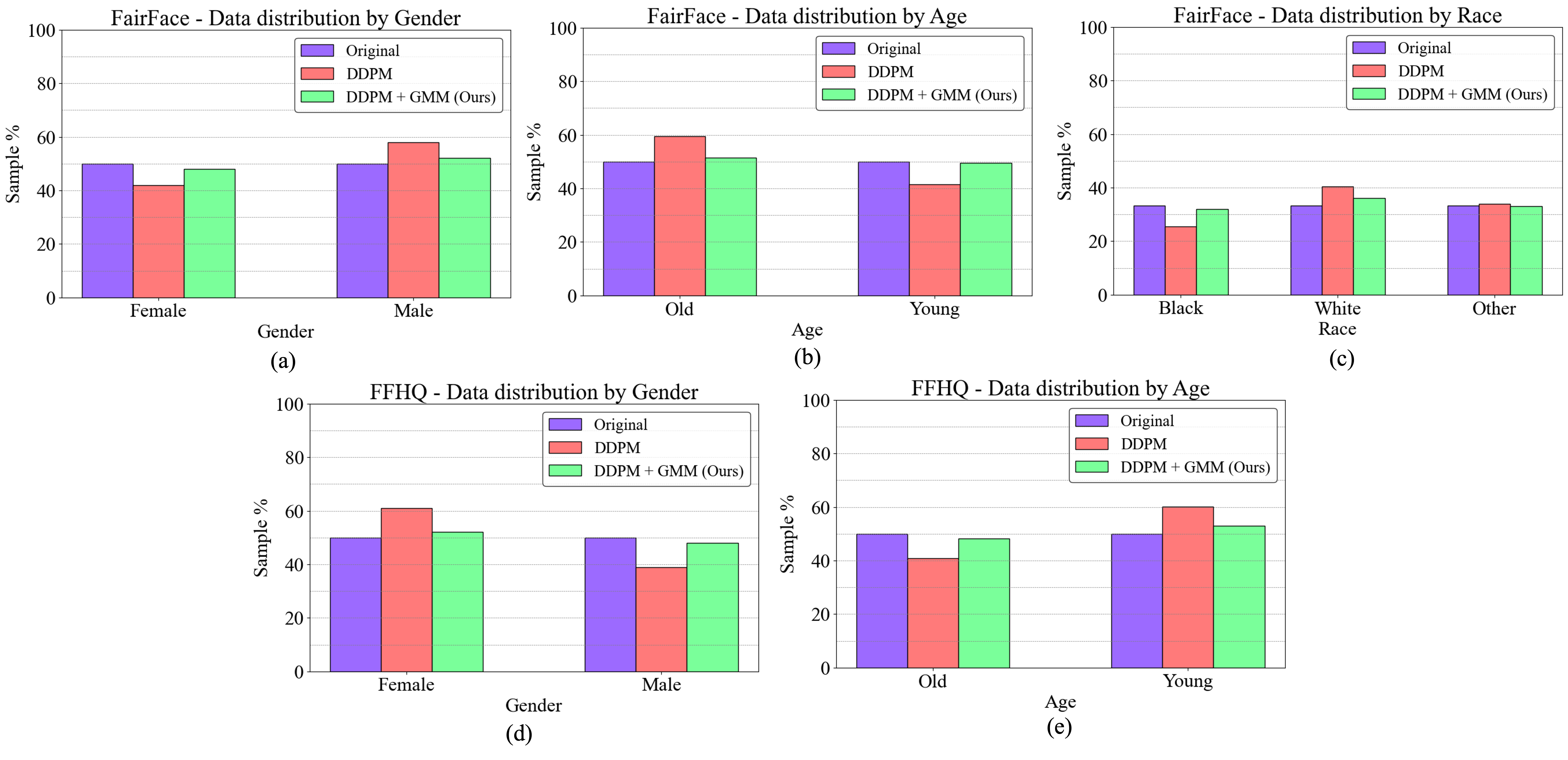}
    \caption{Results of our experiments on balanced data. The original distribution (in purple) of each class of attribute in percentage can be seen as exactly 50\% in age and gender and 33.33\% in race. When the diffusion model is trained on the original balanced data, the unconditionally generated images still show an amplification in the bias for a particular class (as seen in red). Using our approach, we demonstrate the reduction in bias in unconditionally generated images that was introduced by the diffusion model (in green).}
    \label{balanced}
\end{figure*}

\section{Ablation study}
\vspace{-5pt}
In the previous experiments that we carried out, the training data for the diffusion model was itself imbalanced in terms of the specific attributes. We utilized our method in the evaluation process and corrected the bias that was introduced by the diffusion model. In this context, it is interesting to explore the effect of balanced training data on the diffusion model. To examine the impact of balanced datasets on racial, age and gender, bias, we randomly sample images from FairFace to create training data subsets that have a balanced racial, age and gender compositions. Thus the balanced training set for FairFace contained 33.33\% White, 33.33\% Black and 33.33\% ``other'' samples for race, 50\% Male and 50\% Female samples for Gender,  and 50\% for Young and 50\% for Old in terms of Age. The same thing was repeated for FFHQ without considering the race labels. 

We observed that balancing data for training did not have much of an impact on the diffusion model. The diffusion model introduced bias in the generated images and disrupted the balance of each class in each attribute. Our results on bias amplification are consistent with those obtained in \cite{r22}, and we aimed to mitigate this bias. Depicted in Figure \ref{balanced}(a), even when trained on this balanced FairFace dataset, the diffusion model yielded a higher percentage of male images compared to the gender distribution in the training data, implying a predisposition towards generating male faces, thus amplifying training data bias. 

Conversely, Figure \ref{balanced}(d) highlights that employing the diffusion model trained on the FFHQ dataset resulted in a bias towards female images, emphasizing the influence of gender distribution in the training dataset on the direction of bias. Our method acts as a corrective measure by localizing means and integrating them into the reverse diffusion process. In the case of FairFace, where the diffusion model exacerbated the bias towards male images, our technique effectively reduced the number of generated male images while augmenting female representation. Similarly, for FFHQ, sampling from separated distributions alleviated the bias of the diffusion model by generating more male images than female images.

Examining Figure \ref{balanced}(b), the age distribution of the data generated by the diffusion model trained on the complete FairFace dataset reveals a lower proportion of faces in age groups above 40 compared to the training distribution. The FairFace-trained diffusion model amplifies the pre-existing bias towards younger faces observed in the training distribution. A parallel trend is discerned in the case of the FFHQ dataset, as depicted in Figure \ref{balanced}(e), indicating a bias towards generating younger face images. Our method serves as a corrective measure, employing a GMM-based approach in the reverse diffusion process to address this age-related bias. Applied to FairFace, where the existing bias in generating younger faces was amplified, our technique demonstrated a reduction in the number of generated younger faces and an increase in the representation of older age groups. Similarly, when applied to FFHQ, our method generated more older faces, mitigating the bias of the diffusion model and resulting in a more balanced generation of age-diverse facial images. This underscores the efficacy of our approach in counteracting and mitigating age-related biases in the generated data. Figure \ref{balanced}(c) illustrates that the diffusion model trained on the FairFace dataset produced a higher proportion of face images belonging to the White racial group, indicating a potential perpetuation of bias towards this specific group. We address this bias using our method, successfully aligning the distribution with the training data and thereby mitigating the bias introduced by the diffusion models.

\section{Conclusion}
\vspace{-5pt}
Accounting for biases in diffusion-based face generation models is essential to alleviate potential undesirable consequences across diverse downstream applications. In this study, we conducted a comprehensive exploration of biases present in diffusion-based face generation models in terms of sensitive attributes- gender, race, and age. We then proposed a novel strategy, “Gaussian Harmony" to counteract these biases. This approach leveraged GMMs to localize the means of latent codes associated with specific class of each attribute. Our approach has an added advantage over performance, that it does not require to be added in the training step. Therefore, without retraining, we can achieve  bias mitigation in unconditional face generation. In the future, it would be interesting to explore conditional diffusion models and focus on the performance of downstream tasks.

{
    \small
    \bibliographystyle{ieeenat_fullname}
    \bibliography{main}

\begin{thebibliography}{46}
\providecommand{\natexlab}[1]{#1}
\providecommand{\url}[1]{\texttt{#1}}
\expandafter\ifx\csname urlstyle\endcsname\relax
  \providecommand{\doi}[1]{doi: #1}\else
  \providecommand{\doi}{doi: \begingroup \urlstyle{rm}\Url}\fi

\bibitem[Albiero et~al.(2020)Albiero, Zhang, and Bowyer]{r37}
V{\'\i}tor Albiero, Kai Zhang, and Kevin~W Bowyer.
\newblock How does gender balance in training data affect face recognition accuracy?
\newblock In \emph{2020 ieee international joint conference on biometrics (ijcb)}, pages 1--10. IEEE, 2020.

\bibitem[Albiero et~al.(2021)Albiero, Zhang, King, and Bowyer]{r34}
V{\'\i}tor Albiero, Kai Zhang, Michael~C King, and Kevin~W Bowyer.
\newblock Gendered differences in face recognition accuracy explained by hairstyles, makeup, and facial morphology.
\newblock \emph{IEEE Transactions on Information Forensics and Security}, 17:\penalty0 127--137, 2021.

\bibitem[Albiero et~al.(2022)Albiero, Bowyer, and King]{r36}
V{\'\i}tor Albiero, Kevin~W Bowyer, and Michael~C King.
\newblock Face regions impact recognition accuracy differently across demographics.
\newblock In \emph{2022 IEEE International Joint Conference on Biometrics (IJCB)}, pages 1--9. IEEE, 2022.

\bibitem[Antoniou et~al.(2017)Antoniou, Storkey, and Edwards]{r8}
Antreas Antoniou, Amos Storkey, and Harrison Edwards.
\newblock Data augmentation generative adversarial networks.
\newblock \emph{arXiv preprint arXiv:1711.04340}, 2017.

\bibitem[Brock et~al.(2018)Brock, Donahue, and Simonyan]{r3}
Andrew Brock, Jeff Donahue, and Karen Simonyan.
\newblock Large scale gan training for high fidelity natural image synthesis.
\newblock \emph{arXiv preprint arXiv:1809.11096}, 2018.

\bibitem[Choi et~al.(2020{\natexlab{a}})Choi, Grover, Singh, Shu, and Ermon]{r12}
Kristy Choi, Aditya Grover, Trisha Singh, Rui Shu, and Stefano Ermon.
\newblock Fair generative modeling via weak supervision.
\newblock In \emph{International Conference on Machine Learning}, pages 1887--1898. PMLR, 2020{\natexlab{a}}.

\bibitem[Choi et~al.(2020{\natexlab{b}})Choi, Grover, Singh, Shu, and Ermon]{r44}
Kristy Choi, Aditya Grover, Trisha Singh, Rui Shu, and Stefano Ermon.
\newblock Fair generative modeling via weak supervision.
\newblock In \emph{International Conference on Machine Learning}, pages 1887--1898. PMLR, 2020{\natexlab{b}}.

\bibitem[Croitoru et~al.(2023)Croitoru, Hondru, Ionescu, and Shah]{r21}
Florinel-Alin Croitoru, Vlad Hondru, Radu~Tudor Ionescu, and Mubarak Shah.
\newblock Diffusion models in vision: A survey.
\newblock \emph{IEEE Transactions on Pattern Analysis and Machine Intelligence}, 2023.

\bibitem[de~Rosa and Papa(2021)]{r7}
Gustavo~H de Rosa and Joao~P Papa.
\newblock A survey on text generation using generative adversarial networks.
\newblock \emph{Pattern Recognition}, 119:\penalty0 108098, 2021.

\bibitem[Dhar et~al.(2020)Dhar, Bansal, Castillo, Gleason, Phillips, and Chellappa]{r45}
Prithviraj Dhar, Ankan Bansal, Carlos~D Castillo, Joshua Gleason, P~Jonathon Phillips, and Rama Chellappa.
\newblock How are attributes expressed in face dcnns?
\newblock In \emph{2020 15th IEEE International Conference on Automatic Face and Gesture Recognition (FG 2020)}, pages 85--92. IEEE, 2020.

\bibitem[Dhar et~al.(2021{\natexlab{a}})Dhar, Gleason, Roy, Castillo, and Chellappa]{r9}
Prithviraj Dhar, Joshua Gleason, Aniket Roy, Carlos~D Castillo, and Rama Chellappa.
\newblock Pass: protected attribute suppression system for mitigating bias in face recognition.
\newblock In \emph{Proceedings of the IEEE/CVF International Conference on Computer Vision}, pages 15087--15096, 2021{\natexlab{a}}.

\bibitem[Dhar et~al.(2021{\natexlab{b}})Dhar, Gleason, Roy, Castillo, Phillips, and Chellappa]{r10}
Prithviraj Dhar, Joshua Gleason, Aniket Roy, Carlos~D Castillo, P~Jonathon Phillips, and Rama Chellappa.
\newblock Distill and de-bias: Mitigating bias in face verification using knowledge distillation.
\newblock \emph{arXiv preprint arXiv:2112.09786}, 2021{\natexlab{b}}.

\bibitem[Dhariwal and Nichol(2021)]{r29}
Prafulla Dhariwal and Alexander Nichol.
\newblock Diffusion models beat gans on image synthesis.
\newblock \emph{Advances in neural information processing systems}, 34:\penalty0 8780--8794, 2021.

\bibitem[Duarte et~al.(2019)Duarte, Roldan, Tubau, Escur, Pascual, Salvador, Mohedano, McGuinness, Torres, and Giro-i Nieto]{r28}
Amanda~Cardoso Duarte, Francisco Roldan, Miquel Tubau, Janna Escur, Santiago Pascual, Amaia Salvador, Eva Mohedano, Kevin McGuinness, Jordi Torres, and Xavier Giro-i Nieto.
\newblock Wav2pix: Speech-conditioned face generation using generative adversarial networks.
\newblock In \emph{ICASSP}, pages 8633--8637, 2019.

\bibitem[Frankel and Vendrow(2020)]{r11}
Eric Frankel and Edward Vendrow.
\newblock Fair generation through prior modification.
\newblock In \emph{32nd Conference on Neural Information Processing Systems (NeurIPS 2018)}, 2020.

\bibitem[Goodfellow et~al.(2020{\natexlab{a}})Goodfellow, Pouget-Abadie, Mirza, Xu, Warde-Farley, Ozair, Courville, and Bengio]{r1}
Ian Goodfellow, Jean Pouget-Abadie, Mehdi Mirza, Bing Xu, David Warde-Farley, Sherjil Ozair, Aaron Courville, and Yoshua Bengio.
\newblock Generative adversarial networks.
\newblock \emph{Communications of the ACM}, 63\penalty0 (11):\penalty0 139--144, 2020{\natexlab{a}}.

\bibitem[Goodfellow et~al.(2020{\natexlab{b}})Goodfellow, Pouget-Abadie, Mirza, Xu, Warde-Farley, Ozair, Courville, and Bengio]{r23}
Ian Goodfellow, Jean Pouget-Abadie, Mehdi Mirza, Bing Xu, David Warde-Farley, Sherjil Ozair, Aaron Courville, and Yoshua Bengio.
\newblock Generative adversarial networks.
\newblock \emph{Communications of the ACM}, 63\penalty0 (11):\penalty0 139--144, 2020{\natexlab{b}}.

\bibitem[Grover et~al.(2019{\natexlab{a}})Grover, Song, Kapoor, Tran, Agarwal, Horvitz, and Ermon]{r18}
Aditya Grover, Jiaming Song, Ashish Kapoor, Kenneth Tran, Alekh Agarwal, Eric~J Horvitz, and Stefano Ermon.
\newblock Bias correction of learned generative models using likelihood-free importance weighting.
\newblock \emph{Advances in neural information processing systems}, 32, 2019{\natexlab{a}}.

\bibitem[Grover et~al.(2019{\natexlab{b}})Grover, Song, Kapoor, Tran, Agarwal, Horvitz, and Ermon]{r43}
Aditya Grover, Jiaming Song, Ashish Kapoor, Kenneth Tran, Alekh Agarwal, Eric~J Horvitz, and Stefano Ermon.
\newblock Bias correction of learned generative models using likelihood-free importance weighting.
\newblock \emph{Advances in neural information processing systems}, 32, 2019{\natexlab{b}}.

\bibitem[Ho et~al.(2020)Ho, Jain, and Abbeel]{r20}
Jonathan Ho, Ajay Jain, and Pieter Abbeel.
\newblock Denoising diffusion probabilistic models.
\newblock \emph{Advances in neural information processing systems}, 33:\penalty0 6840--6851, 2020.

\bibitem[Huang et~al.(2023)Huang, Chan, Jiang, and Liu]{r32}
Ziqi Huang, Kelvin~CK Chan, Yuming Jiang, and Ziwei Liu.
\newblock Collaborative diffusion for multi-modal face generation and editing.
\newblock In \emph{Proceedings of the IEEE/CVF Conference on Computer Vision and Pattern Recognition}, pages 6080--6090, 2023.

\bibitem[Humayun et~al.(2021)Humayun, Balestriero, and Baraniuk]{r13}
Ahmed~Imtiaz Humayun, Randall Balestriero, and Richard Baraniuk.
\newblock Magnet: Uniform sampling from deep generative network manifolds without retraining.
\newblock In \emph{International Conference on Learning Representations}, 2021.

\bibitem[Hutchinson and Mitchell(2019)]{r14}
Ben Hutchinson and Margaret Mitchell.
\newblock 50 years of test (un) fairness: Lessons for machine learning.
\newblock In \emph{Proceedings of the conference on fairness, accountability, and transparency}, pages 49--58, 2019.

\bibitem[Jain et~al.(2022)Jain, Olmo, Sengupta, Manikonda, and Kambhampati]{r38}
Niharika Jain, Alberto Olmo, Sailik Sengupta, Lydia Manikonda, and Subbarao Kambhampati.
\newblock Imperfect imaganation: Implications of gans exacerbating biases on facial data augmentation and snapchat face lenses.
\newblock \emph{Artificial Intelligence}, 304:\penalty0 103652, 2022.

\bibitem[Jalan et~al.(2020)Jalan, Maurya, Corda, Dsouza, and Panchal]{r16}
Harsh~Jaykumar Jalan, Gautam Maurya, Canute Corda, Sunny Dsouza, and Dakshata Panchal.
\newblock Suspect face generation.
\newblock In \emph{2020 3rd International Conference on Communication System, Computing and IT Applications (CSCITA)}, pages 73--78. IEEE, 2020.

\bibitem[Karkkainen and Joo(2021)]{r46}
Kimmo Karkkainen and Jungseock Joo.
\newblock Fairface: Face attribute dataset for balanced race, gender, and age for bias measurement and mitigation.
\newblock In \emph{Proceedings of the IEEE/CVF winter conference on applications of computer vision}, pages 1548--1558, 2021.

\bibitem[Karras et~al.(2017)Karras, Aila, Laine, and Lehtinen]{r27}
Tero Karras, Timo Aila, Samuli Laine, and Jaakko Lehtinen.
\newblock Progressive growing of gans for improved quality, stability, and variation.
\newblock \emph{arXiv preprint arXiv:1710.10196}, 2017.

\bibitem[Karras et~al.(2019)Karras, Laine, and Aila]{r47}
Tero Karras, Samuli Laine, and Timo Aila.
\newblock A style-based generator architecture for generative adversarial networks.
\newblock In \emph{Proceedings of the IEEE/CVF conference on computer vision and pattern recognition}, pages 4401--4410, 2019.

\bibitem[Karras et~al.(2020)Karras, Laine, Aittala, Hellsten, Lehtinen, and Aila]{r2}
Tero Karras, Samuli Laine, Miika Aittala, Janne Hellsten, Jaakko Lehtinen, and Timo Aila.
\newblock Analyzing and improving the image quality of stylegan.
\newblock In \emph{Proceedings of the IEEE/CVF conference on computer vision and pattern recognition}, pages 8110--8119, 2020.

\bibitem[Krishnapriya et~al.(2020)Krishnapriya, Albiero, Vangara, King, and Bowyer]{r33}
KS Krishnapriya, V{\'\i}tor Albiero, Kushal Vangara, Michael~C King, and Kevin~W Bowyer.
\newblock Issues related to face recognition accuracy varying based on race and skin tone.
\newblock \emph{IEEE Transactions on Technology and Society}, 1\penalty0 (1):\penalty0 8--20, 2020.

\bibitem[Liu et~al.(2021)Liu, Huang, Yu, Wang, and Mallya]{r24}
Ming-Yu Liu, Xun Huang, Jiahui Yu, Ting-Chun Wang, and Arun Mallya.
\newblock Generative adversarial networks for image and video synthesis: Algorithms and applications.
\newblock \emph{Proceedings of the IEEE}, 109\penalty0 (5):\penalty0 839--862, 2021.

\bibitem[Luccioni et~al.(2023)Luccioni, Akiki, Mitchell, and Jernite]{r40}
Alexandra~Sasha Luccioni, Christopher Akiki, Margaret Mitchell, and Yacine Jernite.
\newblock Stable bias: Analyzing societal representations in diffusion models.
\newblock \emph{arXiv preprint arXiv:2303.11408}, 2023.

\bibitem[Maluleke et~al.(2022)Maluleke, Thakkar, Brooks, Weber, Darrell, Efros, Kanazawa, and Guillory]{r39}
Vongani~H Maluleke, Neerja Thakkar, Tim Brooks, Ethan Weber, Trevor Darrell, Alexei~A Efros, Angjoo Kanazawa, and Devin Guillory.
\newblock Studying bias in gans through the lens of race.
\newblock In \emph{European Conference on Computer Vision}, pages 344--360. Springer, 2022.

\bibitem[M{\"u}ller-Franzes et~al.(2023)M{\"u}ller-Franzes, Niehues, Khader, Arasteh, Haarburger, Kuhl, Wang, Han, Nolte, Nebelung, et~al.]{r30}
Gustav M{\"u}ller-Franzes, Jan~Moritz Niehues, Firas Khader, Soroosh~Tayebi Arasteh, Christoph Haarburger, Christiane Kuhl, Tianci Wang, Tianyu Han, Teresa Nolte, Sven Nebelung, et~al.
\newblock A multimodal comparison of latent denoising diffusion probabilistic models and generative adversarial networks for medical image synthesis.
\newblock \emph{Scientific Reports}, 13\penalty0 (1):\penalty0 12098, 2023.

\bibitem[Naik and Nushi(2023)]{r41}
Ranjita Naik and Besmira Nushi.
\newblock Social biases through the text-to-image generation lens.
\newblock \emph{arXiv preprint arXiv:2304.06034}, 2023.

\bibitem[Ojha et~al.(2021)Ojha, Li, Lu, Efros, Lee, Shechtman, and Zhang]{r4}
Utkarsh Ojha, Yijun Li, Jingwan Lu, Alexei~A Efros, Yong~Jae Lee, Eli Shechtman, and Richard Zhang.
\newblock Few-shot image generation via cross-domain correspondence.
\newblock In \emph{Proceedings of the IEEE/CVF Conference on Computer Vision and Pattern Recognition}, pages 10743--10752, 2021.

\bibitem[Perera and Patel(2023)]{r22}
Malsha~V Perera and Vishal~M Patel.
\newblock Analyzing bias in diffusion-based face generation models.
\newblock \emph{arXiv preprint arXiv:2305.06402}, 2023.

\bibitem[Sauer et~al.(2022)Sauer, Schwarz, and Geiger]{r25}
Axel Sauer, Katja Schwarz, and Andreas Geiger.
\newblock Stylegan-xl: Scaling stylegan to large diverse datasets.
\newblock In \emph{ACM SIGGRAPH 2022 conference proceedings}, pages 1--10, 2022.

\bibitem[Singh et~al.(2023)Singh, Gould, and Zheng]{r6}
Jaskirat Singh, Stephen Gould, and Liang Zheng.
\newblock High-fidelity guided image synthesis with latent diffusion models.
\newblock In \emph{2023 IEEE/CVF Conference on Computer Vision and Pattern Recognition (CVPR)}, pages 5997--6006. IEEE, 2023.

\bibitem[Stypu{\l}kowski et~al.(2023)Stypu{\l}kowski, Vougioukas, He, Zi{\k{e}}ba, Petridis, and Pantic]{r31}
Micha{\l} Stypu{\l}kowski, Konstantinos Vougioukas, Sen He, Maciej Zi{\k{e}}ba, Stavros Petridis, and Maja Pantic.
\newblock Diffused heads: Diffusion models beat gans on talking-face generation.
\newblock \emph{arXiv preprint arXiv:2301.03396}, 2023.

\bibitem[Tan et~al.(2020)Tan, Shen, and Zhou]{r19}
Shuhan Tan, Yujun Shen, and Bolei Zhou.
\newblock Improving the fairness of deep generative models without retraining.
\newblock \emph{arXiv preprint arXiv:2012.04842}, 2020.

\bibitem[Teo et~al.(2023)Teo, Abdollahzadeh, and Cheung]{r15}
Christopher~TH Teo, Milad Abdollahzadeh, and Ngai-Man Cheung.
\newblock Fair generative models via transfer learning.
\newblock In \emph{Proceedings of the AAAI Conference on Artificial Intelligence}, pages 2429--2437, 2023.

\bibitem[Wu et~al.(2023)Wu, Albiero, Krishnapriya, King, and Bowyer]{r35}
Haiyu Wu, V{\'\i}tor Albiero, KS Krishnapriya, Michael~C King, and Kevin~W Bowyer.
\newblock Face recognition accuracy across demographics: Shining a light into the problem.
\newblock In \emph{Proceedings of the IEEE/CVF Conference on Computer Vision and Pattern Recognition}, pages 1041--1050, 2023.

\bibitem[Yang et~al.(2022)Yang, Zhang, Song, Hong, Xu, Zhao, Zhang, Cui, and Yang]{r5}
Ling Yang, Zhilong Zhang, Yang Song, Shenda Hong, Runsheng Xu, Yue Zhao, Wentao Zhang, Bin Cui, and Ming-Hsuan Yang.
\newblock Diffusion models: A comprehensive survey of methods and applications.
\newblock \emph{ACM Computing Surveys}, 2022.

\bibitem[Yoon et~al.(2019)Yoon, Hamarneh, and Garbi]{r17}
Chris Yoon, Ghassan Hamarneh, and Rafeef Garbi.
\newblock Generalizable feature learning in the presence of data bias and domain class imbalance with application to skin lesion classification.
\newblock In \emph{Medical Image Computing and Computer Assisted Intervention--MICCAI 2019: 22nd International Conference, Shenzhen, China, October 13--17, 2019, Proceedings, Part IV 22}, pages 365--373. Springer, 2019.

\bibitem[Zhu et~al.(2017)Zhu, Park, Isola, and Efros]{r26}
Jun-Yan Zhu, Taesung Park, Phillip Isola, and Alexei~A Efros.
\newblock Unpaired image-to-image translation using cycle-consistent adversarial networks.
\newblock In \emph{Proceedings of the IEEE international conference on computer vision}, pages 2223--2232, 2017.

\end{thebibliography}
}


\end{document}